\documentclass[12pt]{iopart}

\pdfoutput=1
\usepackage[utf8]{inputenc}
\usepackage[USenglish]{babel}

\usepackage{fancyhdr}
\usepackage{microtype}
\usepackage{graphicx}
\usepackage{subfigure}
\usepackage{booktabs}
\usepackage{amssymb,amsthm,amsfonts}%amsmath
\usepackage{MnSymbol}
\usepackage{mathrsfs}
\usepackage{slashed}
\usepackage{color,rotating}
\usepackage{dsfont}
\usepackage{setspace}
\usepackage{physics}
\usepackage{verbatim}
\usepackage{hyperref,doi}
\usepackage[export]{adjustbox}
\usepackage[normalem]{ulem}
\usepackage{bbold}
\usepackage[linesnumbered,ruled,vlined]{algorithm2e}
\SetKwInput{KwInput}{Input}    
\SetKwInput{KwOutput}{Output}

\usepackage{graphicx}

\usepackage[labelfont=bf, font=footnotesize]{caption}

\makeatletter
\def\l@subsubsection#1#2{}
\makeatother

\newcommand{\beq}{\begin{equation}}
\newcommand{\eeq}{\end{equation}}
\newcommand{\beqa}{\begin{eqnarray}}
\newcommand{\eeqa}{\end{eqnarray}}
\newcommand{\bfc}{\begin{figure}[t]\begin{center}}
\newcommand{\efc}{\end{center}\end{figure}}

\def\Fig#1{Fig.~\ref{#1}}
\def\fig#1{Fig.~\ref{#1}}
\def\tab#1{Table~\ref{#1}}

\def\eq#1{(\ref{#1})}

\def\0#1#2{\frac{#1}{#2}}  %% fractions

\graphicspath{{./figures/}}
\newcommand{\be}{\begin{eqnarray}}
\newcommand{\ee}{\end{eqnarray}}

\begin{document}

\title[How to get the most out of Twinned Regression Methods]{How to get the most out of Twinned Regression Methods}

\author{Sebastian J. Wetzel}
\address{University of Waterloo, Waterloo, Ontario N2L 3G1, Canada}
\address{Perimeter Institute for Theoretical Physics, Waterloo, Ontario N2L 2Y5, Canada}
\address{Homes Plus Magazine Inc., Waterloo, Ontario N2V 2B1, Canada}

%%%%%%%%%%%%%%%%%%%%%%%%%%%%%%%%

\begin{abstract}
Twinned regression methods are designed to solve the dual problem to the original regression problem, predicting differences between regression targets rather then the targets themselves. A solution to the original regression problem can be obtained by
ensembling predicted differences between the targets of an unknown data point and multiple known anchor data points. We explore different aspects of twinned regression methods: 
(1) We decompose different steps in twinned regression algorithms and examine their contributions to the final performance,
(2) We examine the intrinsic ensemble quality,
(3) We combine twin neural network regression with k-nearest neighbor regression to design a more accurate and efficient regression method, and
(4) we develop a simplified semi-supervised regression scheme.
\end{abstract}
\vspace{2pc}
\maketitle

%%%%%%%%%%%%%%%%%%%%%%%%%%%%%%%% 

\section{Introduction}
Regression is one of the most general and common machine-learning tasks, practitioners in many different fields of science and industry rely on methods that help them to make the most accurate and reliable predictions on new data points inferred from a limited amount of training data. Conventional regression methods aim to infer the mapping of input features to one or multiple target variables. Twinned regression methods aim to solve the dual problem of predicting the difference between target values, a solution to the original regression problem can then be obtained by evaluating the predictions between a new unknown data point and multiple anchor data points. This process creates an ensemble of predictions from a single trained model, which tends to be more accurate than solving the original regression problem directly and opens up semi-supervised learning and uncertainty estimates for a very low cost \cite{wetzel2020twin,wetzel2021twin,tynes2021pairwise}. However, the trade-off is the poor scaling with large data sets, since the effective training data set size scales quadratically with the size of the original training data set.

Thus, these methods are beneficial in domains where data is either scarce or costly to obtain. This is the case in for example real estate where markets differ from city to city and data becomes outdated quickly \cite{baldominos2018identifying,rafiei2016novel,yu2021research}. Another example stems from calculations in chemistry where simulations of chemical systems based on quantum mechanics require an enormous amount of computational resources \cite{ryczko2022toward,avula2022building}.

This article is meant to be a practitioner's guide to using \emph{twinned regression} methods that guides the reader through advantages and trade-offs and attempts to answer most questions that were left open in the recent years. 

The main question on our mind is why can such a simple trick of solving the dual problem yield a more accurate prediction than solving the regression problem directly. While we do not fully answer this question, we decomposed twin neural network regression (TNNR) into different steps each with the potential to enhance the performance over traditional algorithms. These include increased effective data set size obtained though pairing training data points, or different ensembles of TNNR predictions. Further, by mapping extreme cases of TNNR to k-nearest neighbor (k-NN) regression and normal artificial neural networks (ANN) we can observe a distinct performance behaviour of twinned regression methods different from traditional regression.

Further, we are eager to improve the accuracy of twinned regression methods. For this purpose, we devise an improvement to TNNR based on the idea of weighting the predictions from different anchors. This leads us to combine k-nearest neighbors (kNN) with TNNR to an even more accurate regression scheme.

The semi-supervised regression framework for TNNR invented in \cite{wetzel2021twin} is specifically tailored to neural network-based regression. It is based on enforcing consistency conditions on unknown data points through a modified loss function. At the end of this manuscript we examine a way to translate a simplified version of this semi-supervised learning scheme to a twinned version of random forests (RF) proposed in \cite{tynes2021pairwise}.

In some projects it is important to apply neural networks with strong memory constraints, this might be the case in small chips or autonomous systems \cite{czischek2021miniaturizing,lechner2018neuronal}. In these cases it would normally be very inefficient to store ensembles of machine learning models due to the increased number of parameters. In contrast to that, with twinned regression methods, one only needs to store additional anchor data points.

\section{Prior Work}

The pairwise comparison inherent to twinned regression methods is inspired by Siamese neural networks which were devised to solve the similarity classification problem as it occurs in fingerprint recognition or signature verification \cite{bromley1993signature,baldi1993neural}. Siamese neural networks contain two identical neural networks with shared weights which project a pair of inputs into a latent space on which the pairwise similarity is determined by the distance. Twinned regression methods also take a pair of inputs to predict the difference between the labels \cite{wetzel2020twin}.

Twin neural network regression \cite{wetzel2020twin} was invented as a regression method that solves the dual problem of predicting pairwise differences between the target values of pairs of input data points. Independently, the same idea has been developed for random forests \cite{tynes2021pairwise}. This kind of regression framework has been shown to have several advantages: (1) it allows for a very efficient generation of ensemble predictions \cite{wetzel2020twin,tynes2021pairwise}. Typically in methods that generate ensembles from training a single machine learning model, the predictions are strongly correlated \cite{srivastava2014dropout,wan2013regularization} since they can be deformed into each other through small perturbations. In twinned regression methods however, ensemble members are separated by the distance of the input data points themselves. (2) Twinned regression methods tend to be more accurate than the underlying base algorithm on many data sets \cite{wetzel2020twin,tynes2021pairwise}, (3) consistency conditions allow for the formulation of uncertainty estimators in addition to the ensemble variance \cite{wetzel2020twin,tynes2021pairwise} and (4) loops containing unlabelled data points can be supplied while training, hence turning the method into a semi-supervised regression algorithm \cite{wetzel2021twin}. Further, (5) the intrinsic uncertainty estimation lends itself for active learning \cite{tynes2021pairwise}.

A central contribution of this article is the combination of twin neural network regression and k-NN regression to increase the accuracy of over standard twin neural network regression. Similarly, artificial neural networks have been employed in tandem with k-NN regression in different contexts before \cite{wu2009novel,bensaci2021deep,liu2018hybrid}.

\section{Reformulation of the Regression Problem}
\begin{figure*}[h!]
    \centering
    \includegraphics[width=0.9\textwidth]{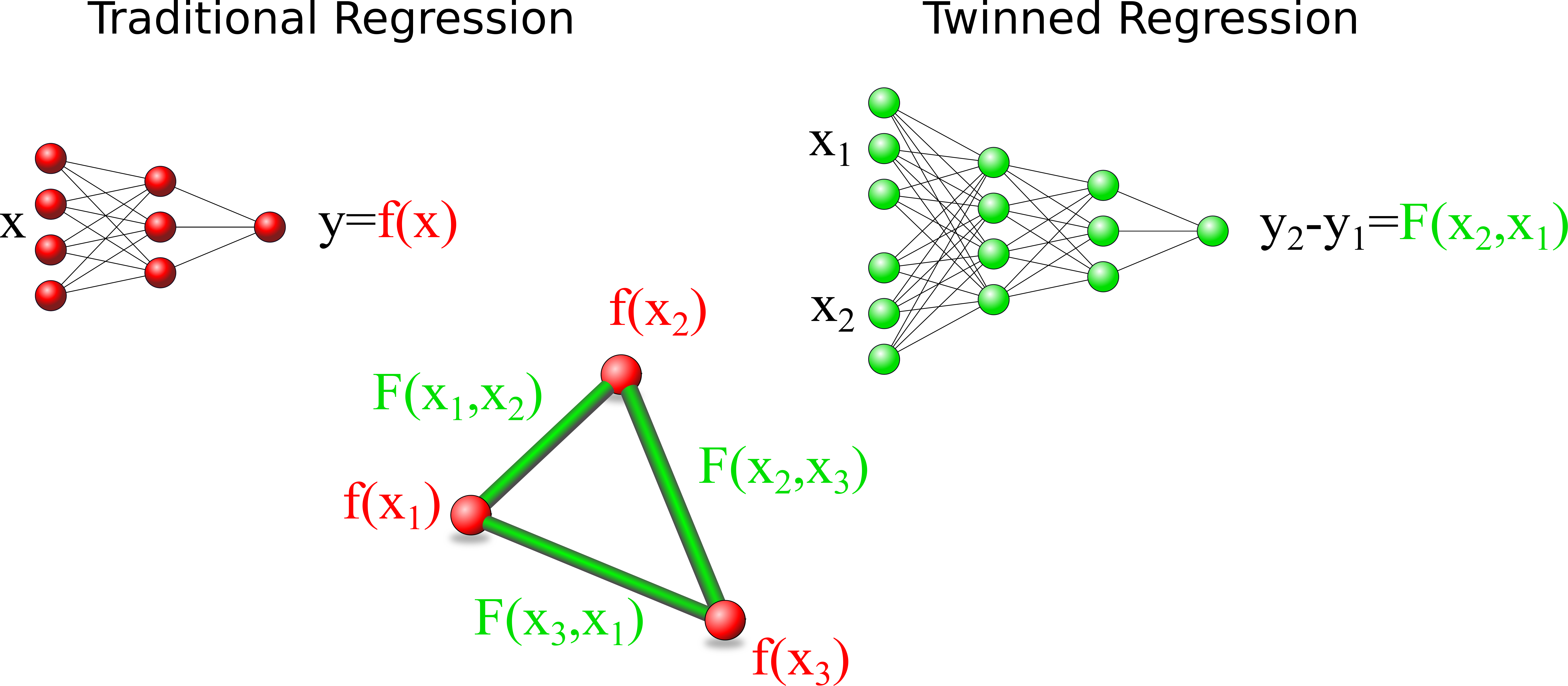}
    \caption{Dual formulation of a regression problem: A traditional solution to a regression problem consists of finding an approximation to the function that maps a data point $x$ to its target value $f(x)=y$. Twinned regression methods solve the dual problem of mapping a pair of inputs $x_1$ and $x_2$ to the difference between the target values $F(x_2,x_1)=y_2-y_1$. The resulting function can then be employed as an estimator for the original regression problem $y_2=F(x_2,x_1)+y_1$ given a labelled anchor point $(x_1,y_1)$. Twinned regression methods must satisfy loop consistency: predictions along each loop sum to zero:  $F(x_1,x_2)+F(x_2,x_3)+F(x_3,x_1)=0$.}
    \label{fig:architecture}
\end{figure*}

A regression problem can be formulated as follows: Given a labelled training data set of $n$ data points $X^{train}=(x_1^{train},...,x_n^{train})$ with their corresponding target values $Y^{train}=(y_1^{train},...,y_m^{train})$, we are tasked to find a function $f$ such that the deviation between $f(x_i)$ and $y_i$ is minimized with respect to a predefined objective function for all data points $x_i$ on the data manifold. In this work, this function is the root mean square error $L_{RMSE}=\sqrt{\sum_{i=1}^n (f(x_i)- y_i)^2}$. Unless stated otherwise, all performance measures are evaluated on unknown test data $(X^{test},Y^{test})$.

Twinned regression methods aim to solve a reformulation of the original regression problem which is visualized in \fig{fig:architecture}. For each pair of data points $(x_i^{train},x_j^{train})$ we train a regression model to find a function $F$ to predict the difference 
\begin{align}
    F(x_i,x_j)= y_i-y_j \quad .    \label{eq:TNNR}
\end{align}

This function $F$ can be used to construct a solution to the original regression problem via $y_i^{pred}= F(x_i,x_j)+y_j$, where $(x_j,y_j)$ is an {\em anchor} whose target value is known. Every training data point $x_j^{train}\in X^{train}$ can be used as such an anchor. A more accurate estimate for the solution of the original regression problem is obtained by averaging over many differences between a fixed unknown data point and different anchor data points
\begin{align}
    y_i^{pred}&=  \frac{1}{n}\sum_{j=1}^n \left(F(x_i,x_j^{train})+y_j^{train}\right)\nonumber \\
    &= \frac{1}{n}\sum_{j=1}^n\left( \frac{1}{2}F(x_i,x_j^{train})-\frac{1}{2}F(x_j^{train},x_i)+y_j^{train} \right) \quad .\label{eq01}
\end{align}
The increase in accuracy is based on averaging out the noise from different anchors and the reduction of the variance error via an ensemble of predictions. Previous works \cite{wetzel2020twin,tynes2021pairwise} recommended using the whole training data set as anchors, hence creating an ensemble of difference predictions $y_i-y_j$ which is twice as large as the training set for every single prediction of $y_i$. 

A major advantage of the dual formulation is the description via loops containing multiple data points as can be seen in \fig{fig:architecture}. In contrast to traditional regression, the results of twinned regression methods need to satisfy consistency conditions, for example for each three data points $x_1,x_2,x_3$, summing up the predictions along a closed loop should yield zero: $F(x_1,x_2)+F(x_2,x_3)+F(x_3,x_1)=0$. During inference, violations of these consistency conditions give rise to uncertainty 
estimates \cite{wetzel2020twin,tynes2021pairwise}. Enforcing loop consistency on predictions involving unlabelled data points in the training phase is what makes twinned regression methods into semi-supervised regression algorithms.

While neural networks are naturally good learners of linear functions this is not the case for other algorithms like random forests. For this reason, \cite{tynes2021pairwise} proposed to augment the input features by their difference $(x_i,x_j)\rightarrow(x_i,x_j,x_i-x_j)$. One might argue that this improvement is similar to common data augmentation, however, this is a feature that traditional machine learning algorithms don't have access to, because it requires two different data points.

\section{Notes About Experiments}
All experiments in this article are performed on the data sets outlined in \ref{sec:data}. Since only neural network based methods scale favorably with the data set size, they use the full data sets of which 70\% are used for training, 10\% as validation set and 20\% as test set. The details of the neural network architectures can be found in the appendix \ref{sec:nn_architecture}. Random forests and especially twinned random forests scale poorly with the data set size thus only 100 training data points are chosen from the data sets and 100 data points comprise the test sets. Random forests do not need validation sets, since the hyper=parameters are optimized via 5-fold cross-validation. In section \ref{sec:rf} one can find the details about our random forest implementations. All experiments are repeated for 25 random but fixed splits of training, test, and if applicable, validation data.

\section{Ensemble Performance}
\begin{figure}
    \centering
    \includegraphics[width=0.95\textwidth]{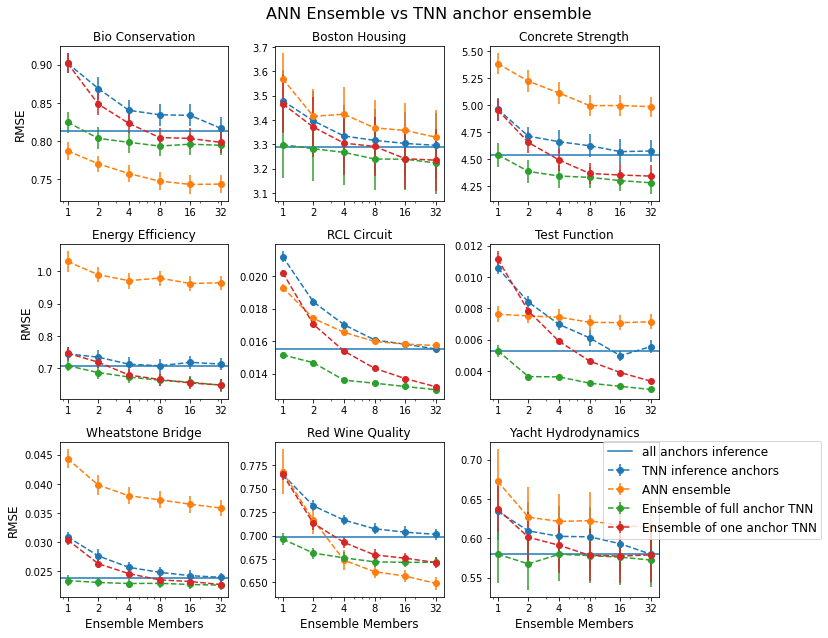}
    \caption{Comparison of different ensembles of ANNs and TNNR measured by RMSE vs number of ensemble members for the dashed lines. Ensembles can be created either by training several models or evaluating TNNR for different anchors. The blue solid line corresponds to training and evaluating one TNNR model on all possible training pairs and predicting the results with all possible anchors. The dashed blue line varies the number of anchors during the inference phase and converges to the solid blue line in the limit of increasing the inference anchors to the full training set. The dashed orange line indicates traditional ANN ensembles where multiple ANNs are trained independently. The dashed green line corresponds to ensembles of independently trained TNNR models, if the ensemble size is 1, this is equivalent to the solid blue line. The dashed red line corresponds to independently trained TNNR models each having only one inference anchor per prediction.}    \label{fig:A}
\end{figure}
\begin{table}
\centering
  \caption{Best estimates for test RMSEs obtained by artificial neural network(ANN) regression compared to ensembles of ANN regression. Our confidence on the RMSEs is determined by their standard error. We train on 70\% of the available data, 10\% validation data, 20\% test data. 
  }
 
  \begin{tabular}{l llr}\\
    %%%%%%%   & \multicolumn{3}{l}{80\% labelled training data}            \\

 & Single ANN & 32 ANN ensemble & Gain \\
     \cmidrule(r){2-2}\cmidrule(r){3-4}
Bio Conservation(BC) & 0.7874$\pm$0.012 & 0.7438$\pm$0.0118 & 5.53\% \\
Boston Housing(BH) & 3.5695$\pm$0.1051 & 3.3291$\pm$0.1137 & 6.73\% \\
Concrete Strength(CS) & 5.3829$\pm$0.0964 & 4.9842$\pm$0.0938 & 7.41\% \\
Energy Efficiency(EE) & 1.0310$\pm$0.0313 & 0.9649$\pm$0.0224 & 6.42\% \\
RCL Circuit(RCL) & 0.0193$\pm$0.0002 & 0.0157$\pm$0.0002 & 18.41\% \\
Test Function(TF) & 0.0076$\pm$0.0005 & 0.0071$\pm$0.0005 & 6.35\% \\
Wheatstone Bridge(WSB) & 0.0443$\pm$0.0016 & 0.0359$\pm$0.0014 & 19.1\% \\
Red Wine Quality(WN) & 0.7681$\pm$0.024 & 0.6487$\pm$0.0068 & 15.54\% \\

Yacht Hydrodynamics(YH) & 0.6723$\pm$0.0406 & 0.6143$\pm$0.0357 & 8.63\%

  \end{tabular}
  \label{tab:ann_ensemble}
\end{table}

\begin{table}
\centering
  \caption{Best estimates for test RMSEs obtained by Twin Neural Network Regression (TNNR). We measure the improvement between using a single anchor during inference phase and using all anchors. Further, the latter is compared to an ensemble of full anchor TNNR.
  }
 
  \begin{tabular}{l llrlr}\\
    %%%%%%%   & \multicolumn{3}{l}{80\% labelled training data}            \\

 & 1 Anchor TNNR & All Anchor TNNR & Gain & Ensemble of 32 TNNR & Gain \\
      \cmidrule(r){2-2}\cmidrule(r){3-4}\cmidrule(r){5-6}
BC & 0.9021$\pm$0.0131 & 0.8140$\pm$0.0149 & 9.77\% & 0.7947$\pm$0.0134 & 2.37\% \\
BH & 3.4793$\pm$0.1262 & 3.2897$\pm$0.1293 & 5.45\% & 3.2232$\pm$0.1256 & 2.02\% \\
CS & 4.9602$\pm$0.1073 & 4.5385$\pm$0.1124 & 8.5\% & 4.2791$\pm$0.0995 & 5.71\% \\
EE & 0.7445$\pm$0.0224 & 0.7071$\pm$0.0229 & 5.02\% & 0.6468$\pm$0.0196 & 8.53\% \\
RCL & 0.0212$\pm$0.0003 & 0.0155$\pm$0.0002 & 26.6\% & 0.0130$\pm$0.0001 & 16.23\% \\
TF & 0.0106$\pm$0.0004 & 0.0053$\pm$0.0004 & 50.24\% & 0.0028$\pm$0.0001 & 46.41\% \\
WSB & 0.0309$\pm$0.0009 & 0.0239$\pm$0.0009 & 22.81\% & 0.0227$\pm$0.0008 & 5.15\% \\
WN & 0.7654$\pm$0.0059 & 0.6985$\pm$0.0064 & 8.75\% & 0.6713$\pm$0.0055 & 3.89\% \\

YH & 0.6344$\pm$0.0363 & 0.5798$\pm$0.0362 & 8.62\% & 0.5723$\pm$0.0336 & 1.29\%

  \end{tabular}
  \label{tab:tnnr_ensemble}
\end{table}

Twinned regression methods have been shown to produce accurate solutions to regression problems \cite{wetzel2020twin,tynes2021pairwise}, comparable to or better than other current state-of-the-art algorithms at the cost of scaling poorly towards larger data sets. This naturally leads to the question of where the increased performance stems from. The reformulation of a regression problem into its dual problem of predicting differences between target values opens up several potential reasons for improved accuracy. These include increased effective training set size, internal ensembling of predictions (see explanation in \ref{app:bv}), or the nature of solving a different problem. In the following we examine these reasons at the example of TNNR, however, we assume the answers will also be valid for other baseline algorithms.

Let us start with discussing different kinds of ensembles and their effect on accuracy. \Fig{fig:A} contains the results of several experiments examining the performance of different ensemble types of ANN regression and TNNR. For each data set, the baseline results are the solid blue horizontal line, which represents the test RMSE after applying standard full anchor TNNR and the leftmost point of the orange line which represents the results of applying a single ANN, confirming that TNNR almost always yields a lower RMSE than ANN regression. 

In order to compare traditional ANNs with TNNR, we observe that after training we can map TNNR to an ANN for each anchor, since both ANN regression and TNNR use the same internal architecture. For each fixed $x_j\in X^{train}$ an ANN $\tilde f$ is defined through
\begin{align}
\tilde f_j (x_i) :=F(x_i,x_j)+y_j
\label{eq:ann_tnnr}
\end{align}
The features of $x_j$ modify the weights of $\tilde f_j$ while $y_j$ is absorbed by the output neuron inside its bias. At that point, the only difference between each $\tilde f_j$ and an equivalent ANN is the procedure with which the weights were optimized.

This gives us access to a framework to directly compare ensembles of ANNs and the implicit ensembles generated by TNNR using multiple anchors during inference. We examine the results of these both models through the orange and blue dashed lines in \fig{fig:A}. While both curves reduce the RMSE as we increase the ensemble size, we come to the sobering conclusion that TNNR ensembles and ANN ensembles are not equivalent, they neither have a uniform slope nor do they converge to similar RMSEs.

We have just used a single trained TNNR model for all ensemble members while training each ANN model from scratch. What happens if we retrain TNNR for each single anchor? The results of these experiments are visualized in the red dashed line. Since retraining increases the ensemble diversity the red line is consistently below the blue line. Further, we can see that the performance of these independently trained TNNRs increases faster with the number of anchors.
If the resources are available one can further create an ensemble of different TNNR models each having access to all anchors depicted in the green line. On seven out of nine data sets this yields clearly the best performance. We note that one-anchor TNNR with retraining (red line) converges towards the green for more than 32 ensemble members. This tells us that the full ensemble diversity through multiple anchors and multiple models can be captured independently.

A more quantitative version of the out-performance of TNNR ensembles can be seen in \tab{tab:ann_ensemble} and \tab{tab:tnnr_ensemble}. The magnitude of the \% improvement of the combined anchor+direct ensembling containing multiple TNNs causes a much larger improvement than the ensembling of traditional ANNs.

\section{Effective Training Set Size}
\begin{figure}
    \centering
    \includegraphics[width=0.95\textwidth]{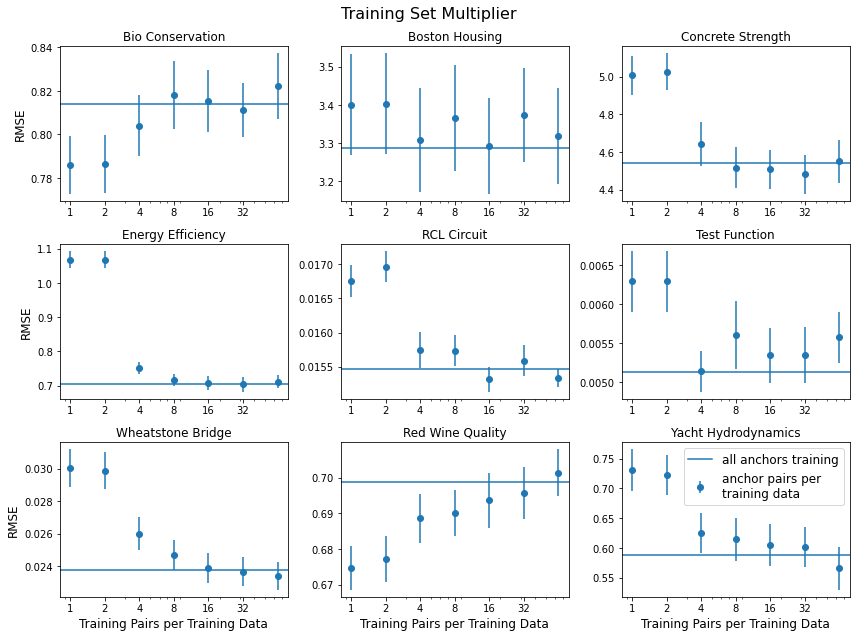}
    \caption{How many different pairings are used for training effects the performance of TNNR measured by RMSE. The blue solid line corresponds to training TNNR on all possible training pairs. It is compared to training TNNR on a randomly chosen but fixed data set of pairs while still employing all training data points as inference anchors. A training set multiplier indicates on how many pairs are taken into account compared to the original unpaired training set. }
    \label{fig:E}
\end{figure}

Another improvement over a traditional regression analysis is the increased training set size that comes from preparing training sets by pairing each training data point with every other training data point. This transforms a training set of size $n$ into a pairwise training set of size $n^2$. In this section, we measure if the increase in the number of pairings leads to an increase in accuracy.

To address this question we look at several curves in \fig{fig:E}. In this figure, we compare the effect of increasing the effective pairwise training data set on the accuracy. For this purpose, we define the training set multiplier as the number of pairs that are created from the original training set to produce the paired training set. A training set multiplier of one means that each training data point is paired with only one other randomly chosen (without replacement) but fixed data point (on average this means each training data point is used twice). Increasing the training set multiplier to the size of the training set converges to the standard formulation of twinned regression methods. We can see that in all data sets, except two, increasing the training set multiplier increases the performance of TNNR. More precisely, a training set multiplier of $\approx 8-16$ seems already to be enough to reach the accuracy of standard TNNR. It is important to note, that on two data sets, namely Bio Conservation(BC) and Red Wine Quality(WN), increasing the training set multiplier has the effect of reducing the performance. This coincides with other algorithms beating TNNR (\fig{fig:A},\fig{fig:B}) and is a sign that TNNR might not be suitable for such regression tasks. A data scientist using TNNR might do a training set multiplier check, if he finds a decreasing accuracy while increasing the multiplier, he can reject TNNR as optimal regression algorithm.

\section{Nearest Neighbor TNNR}
\begin{figure}
    \centering
    \includegraphics[width=0.95\textwidth]{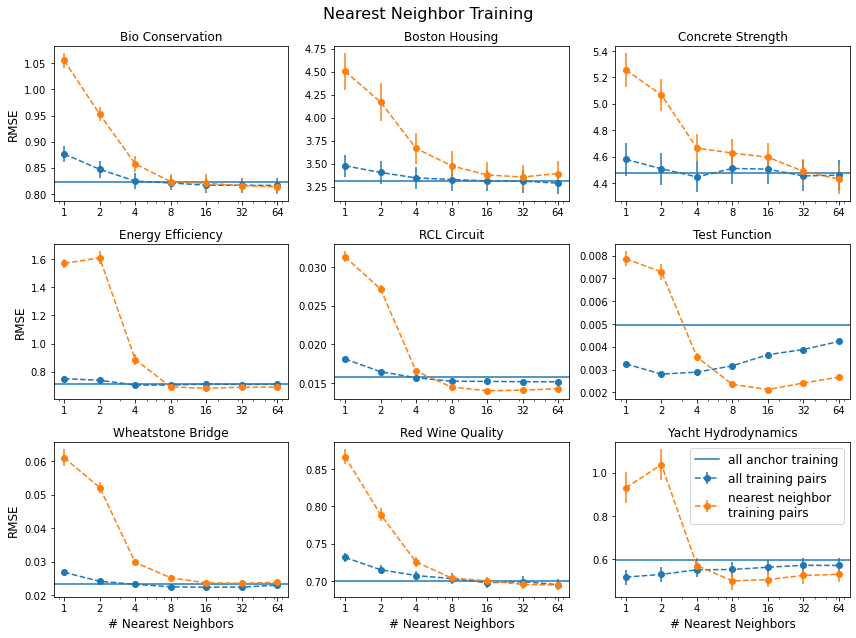}
    \caption{Effect of nearest neighbor pairing on TNNR measured in terms of RMSE vs the number of nearest neighbors used if applicable. The solid blue line marks the performance of the original TNNR. The dashed blue line displays the results of TNNR trained on all possible pairs while performing inference only using nearest neighbor anchors. The dashed orange line is produced by restricting pairs to nearest neighbors for training and inference.}
    \label{fig:D}
\end{figure}

\begin{table}
\centering
  \caption{Best estimates for test RMSEs obtained by Nearest Neighbor Twin Neural Network Regression (NNTNNR). 
  }
 
  \begin{tabular}{l llrlr}\\
    %%%%%%%   & \multicolumn{3}{l}{80\% labelled training data}            \\

 & TNNR & NN inference & Gain & NN train+inference & Gain \\
       \cmidrule(r){2-2}\cmidrule(r){3-4}\cmidrule(r){5-6}

BC & 0.8234$\pm$0.0144 & 0.8162$\pm$0.0155 & 0.87\% & 0.8133$\pm$0.0152 & 1.23\% \\
BH & 3.3104$\pm$0.1202 & 3.2898$\pm$0.1202 & 0.62\% & 3.3563$\pm$0.1161 & -1.39\% \\
CS & 4.4731$\pm$0.1242 & 4.4458$\pm$0.1091 & 0.61\% & 4.4290$\pm$0.1174 & 0.99\% \\
EE & 0.7156$\pm$0.0204 & 0.7056$\pm$0.0193 & 1.4\% & 0.6825$\pm$0.0216 & 4.63\% \\
RCL & 0.0158$\pm$0.0002 & 0.0151$\pm$0.0003 & 3.98\% & 0.0140$\pm$0.0002 & 11.33\% \\
TF & 0.0050$\pm$0.0001 & 0.0028$\pm$0.0002 & 43.65\% & 0.0021$\pm$0.0003 & 57.26\% \\

WSB & 0.0233$\pm$0.0006 & 0.0224$\pm$0.0009 & 4.06\% & 0.0236$\pm$0.0009 & -1.01\% \\
WN & 0.6998$\pm$0.006 & 0.6951$\pm$0.0062 & 0.68\% & 0.6944$\pm$0.006 & 0.77\% \\
YH & 0.5977$\pm$0.0344 & 0.5184$\pm$0.0331 & 13.26\% & 0.5009$\pm$0.0333 & 16.19\%

  \end{tabular}
  \label{tab:knn_tnnr}
\end{table}
In this manuscript we propose a new regression algorithm based on a combination of k-nearest neighbor regression and TNNR, which of course could be implemented for various baseline twinned regression algorithms. In standard twinned regression methods the model learns to predict differences between the targets of two arbitrary data points. This model is then employed to create an ensemble prediction via averaging the approximations of the differences between the target value of a new data point and all anchor data points, see \eq{eq01}. However, not all of these anchor data points might be of equal importance for the prediction. That is why in this section we restrict the anchor points to the nearest neighbors. For this purpose, we define the notation $\text{NN}(i,m)$ as the set of $m$ nearest neighbors of a data point $x_i \in X$ within the training set $x_i \in X^{train}$ to reformulate the prediction:

\begin{align}
    y_i^{pred}&=  \frac{1}{m}\sum_{j\in \text{NN}(i,m)} \left( F(x_i,x_j^{train})+y_j^{train} \right)
\end{align}

While we have defined the prediction using nearest neighbors during the inference phase, it is an open question whether it is better to train the model to predict differences between target values of generic data points or between neighboring data points corresponding to the same number of nearest neighbors in the inference phase. The different training versions are compared in \fig{fig:D} where the baseline is set by standard TNNR. We emphasize that both versions of training obey the same principle for selecting nearest neighbors during inference. When only using the very nearest neighbor as an anchor for inference, we can see that for 7 out of 9 data sets both training versions underperform traditional TNNR, while training on all possible pairs performs better than just training on neighboring training data points. This picture changes as we increase the number of nearest neighbors. On all data sets both versions of nearest neighbor TNNR converge to standard TNNR in the limit of increasing the number of neighbors to the training set size. In 7 out of 9 data sets there is a sweet spot where nearest neighbor TNNR with nearest neighbor training outperforms at around 16 to 64 neighbors, in 3 out of those data sets nearest neighbor training outperforms by a very large margin culminating in reducing the RMSE by $\approx 60\% $ on the TF data set, see \tab{tab:knn_tnnr}. We note that this is the data set with zero noise.

\section{TNNR vs k-NN}
\begin{figure}
    \centering
    \includegraphics[width=0.95\textwidth]{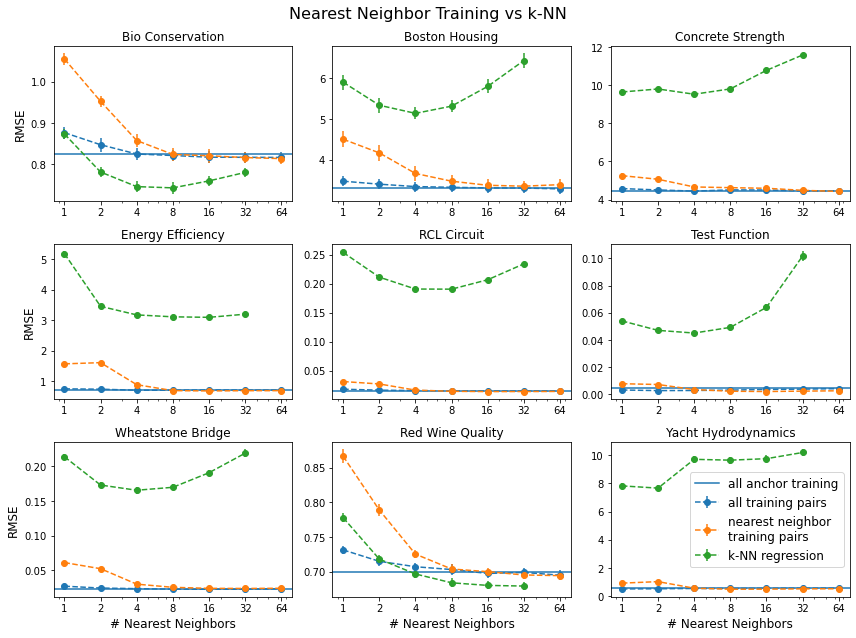}
    \caption{Comparison of k-NN regression and TNNR with different numbers of nearest neighbor training pairs measured by RMSE vs the number of neighbors for the dashed lines. The solid blue corresponds to normal TNNR with access to all possible pairs during training and inference. The blue dashed line restricts the pairs to be nearest neighbors during inference. The orange dashed line restricts the pairs to be nearest neighbors during training and inference. The green dashed line describes k-NN.}
    \label{fig:B}
\end{figure}

A natural question is how nearest-neighbor TNNR(NNTNNR) related to k-NN regression. Nearest neighbor TNNR can be related to k-NN regression through setting $F(x_i,x_j)\equiv 0$, then 

\begin{align}
    y_i^{pred}&=  \frac{1}{m}\sum_{j\in NN(i,m)} \left( \underbrace{F(x_i,x_j^{train})}_{0}+y_j^{train} \right)=\frac{1}{m}\sum_{j\in \text{NN}(i)}^m y_j^{train}
    \label{eq:nntnnr_knn}
\end{align}
Assuming $F(x_i,x_j)$ would just be a minor contribution to k-NN regression we would see a qualitatively similar performance of NNTNNR. In order to test this statement, we visualize the behavior of k-NN and NNTNNR in \fig{fig:B}. In this figure we can clearly see, that NNTNNR beats k-NN regression by an enormous margin on 7 out of 9 data sets. However, there are two data sets, namely BC, WN where k-NN is the winner. Again, we note that these data sets are exactly where the expected TNNR mechanism fails \fig{fig:E} and it coincides with ANN ensembles outperforming TNNR \fig{fig:A}. Further, the number of optimal TNNR anchors is much larger than the optimal number of neighbors in k-NN. 

\section{Reformulation Benefits}

After having discussed the impacts of ensembling and the increased effective training set size, we have now finally the tools to partially answer the question of whether the reformulation to the dual problem itself contributes to the increased accuracy of twinned regression methods. We have related 
TNNR to normal ANNs in \eq{eq:ann_tnnr} and connected NNTNNR to k-NN regression in \eq{eq:nntnnr_knn}. If TNNR would be a glorified form of ANN or k-NN regression, the performance of TNNR could be related qualitatively to neural networks or k-NN. However, as we can see by comparing with ANNs in \fig{fig:A} or k-NN \fig{fig:B}, it is clear that TNNR has a distinct performance profile that beats ANNs and k-NN on the same 7 of 9 data sets and underperforms both on the remaining two data sets. As we know from testing the impact of the increased training set size \fig{fig:E}, these are the data sets where increasing the data sets has an adverse effect on accuracy, which signals that the TNNR mechanism fails while ANN and k-NN continue to perform normally. All these facts support the conclusion that the reformulation to a dual problem itself tends to have a positive effect on accuracy on most data sets.

\section{Miniaturizing accurate networks}
\begin{figure}
    \centering
    \includegraphics[width=0.95\textwidth]{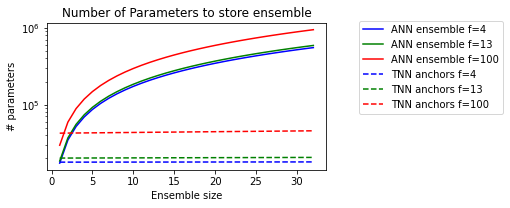}
    \caption{Memory requirements for storing real ensembles in terms of number of parameters stored vs the ensemble size. The architecture used corresponds to the architecture that was used to produce the results in this manuscript. Different architectures cause quantitatively different plots, but qualitatively the behave similarly. The solid lines indicate the memory requirements for storing an ensemble of independently trained ANN models for different numbers of features $f$ describing each data instance. The dashed lines correspond to ensembles generated by storing one TNNR model and a representative number of anchors which can be combined to produce ensembles.}
    \label{fig:F}
\end{figure}
Often it is required to store fully trained neural networks on hardware that has strict memory limitations. This might be on chips that allow for autotuning of quantum dots \cite{czischek2021miniaturizing} or in self driving vehicles \cite{lechner2018neuronal}. In these cases it is required to consider the trade-off between accuracy and memory requirements. Ensembles of neural networks tend to be more accurate than single neural networks, however, storing them requires linearly more memory capacity per each ensemble member. TNNR provides an elegant solution to this problem, because in order to store an ensemble of predictions it only requires the storage of a single set of trained weights and biases together with one anchor data point per ensemble member. Our neural network architectures are chosen such that they produce accurate results on all nine considered data sets. Thus, we have chosen an architecture with two hidden layers, each with 128 neurons. In \fig{fig:F} we visualize the number of parameters that are required to be stored in the case of traditional ANN and TNNR ensembles for different feature sizes of the input data. It can be seen that in all cases TNNR parameters plus anchors need far less storage capacity to achieve a similar ensemble size as ANNs. Of course in practice the optimal neural network architecture and its number of parameters varies between problems, meaning our quantitative analysis of memory requirements might not generalize to other problems. However, the qualitative trend remains the same as long as the feature size is smaller than the number of parameters of the model.

\section{Making other models semi-supervised}
\begin{figure}
    \centering
    \includegraphics[width=0.95\textwidth]{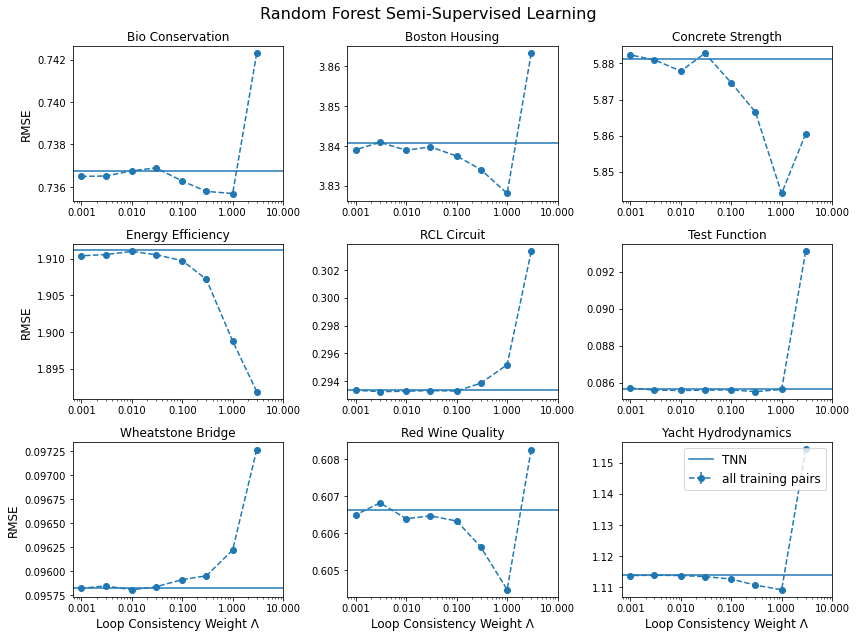}
    \caption{Transductive semi-supervised learning with random forests. Random forests have been supplied with loops containing unlabelled data points. The magnitude of the influence the loops have on the training process is measured by $\Lambda$. Tuning $\Lambda\approx1$ reduces the RMSE on almost all data sets compared to purely supervised learning (solid blue line). }
    \label{fig:G}
\end{figure}
\begin{table}
\centering
  \caption{Best estimates for test RMSEs obtained by Semi-supervised random forests.
  }
 
  \begin{tabular}{l llr}\\
    %%%%%%%   & \multicolumn{3}{l}{80\% labelled training data}            \\

 & Supervised RF & Semi-Supervised RF & Improvement \\
  \cmidrule(r){2-2}\cmidrule(r){3-4}
Bio Conservation & 0.7367$\pm$0.0081 & 0.7357$\pm$0.0078 & 0.14\% \\
Boston Housing & 3.8408$\pm$0.1322 & 3.8281$\pm$0.1356 & 0.33\% \\
Concrete Strength & 5.8813$\pm$0.2242 & 5.8440$\pm$0.2276 & 0.63\% \\
Energy Efficiency & 1.9112$\pm$0.0417 & 1.8919$\pm$0.0411 & 1.01\% \\
RCL Circuit & 0.2934$\pm$0.0068 & 0.2932$\pm$0.0068 & 0.05\% \\
Test Function & 0.0857$\pm$0.0022 & 0.0855$\pm$0.0021 & 0.19\% \\
Wheatstone Bridge & 0.0958$\pm$0.0025 & 0.0958$\pm$0.0022 & 0.02\% \\
Red Wine Quality & 0.6066$\pm$0.0083 & 0.6045$\pm$0.0084 & 0.35\% \\
Yacht Hydrodynamics & 1.1138$\pm$0.0383 & 1.1091$\pm$0.0384 & 0.42\%

  \end{tabular}
  \label{tab:rf_ssl}
\end{table}
In this section we explore a simple framework to train any twinned regression method in a semi-supervised manner. The idea is based on the semi-supervised regression method devised in \cite{wetzel2021twin} for TNNR. As we can see in \fig{fig:architecture} the dual formulation requires machine learning models to predict differences $F(x_i,x_j)=y_i-y_h$ between target values instead of the targets $f(x)=y$, themselves. One advantage of this formulation is that a correct solution would satisfy loop consistency $F(x_1,x_2)+F(x_2,x_3)+F(x_3,x_1)=0$.

Hence, we propose the following algorithm that is applicable to all twinned regression methods: At first we train the regression model on the labelled training data. This model is then used to predict the differences between targets along loops randomly sampled from an unlabelled data set. For each loop an adjustment $a$ defined by
\begin{align}
a=F(x_1,x_2)+F(x_2,x_3)+F(x_3,x_1)
\end{align}
is then used to propose a label $y_{ij}=F(x_i,x_j)-\Lambda \times a $ for each combination of $x_i,x_j$ within the loops. Here, $\Lambda$ is the loop weight hyper-parameter. The unlabelled data set together with the proposed labels is then added to the labelled training set, on which the model is retrained. The algorithm is further depicted in \fig{alg:semi_sup}.

We apply this idea to the pairwise/twinned random forest regression proposed in \cite{tynes2021pairwise}, which was originally aimed at solving regression problems on small data sets in chemistry. Since random forests don't scale as well with large data sets, we restrict our data sets to 100 training and 100 test data points. The details of the training process are outlined in section \ref{sec:rf}.

Before applying the semi-supervised learning strategy, we convince ourselves that twinned random forest regression is suitable for the test bed consisting of the nine data sets (\ref{sec:data}) used in this paper. The corresponding results can be seen in \tab{tab:twin_rf}. Twinned random forests perform equally well, or slightly worse, compared to traditional random forests on three data sets (BC,EE,WN). It moderately outperforms on four data sets (BH,CS,RCL,YH) and it massively outperforms by cutting the RMSE by more than 35\% on two data sets (TF,WSB). 

After having convinced ourselves of the superior performance of twinned random forests, we apply our semi-supervised learning framework in a transductive manner. Transductive means that the test data is used as unlabelled training data. This is in contrast to inductive semi-supervised learning where the unlabelled training data would be kept separate from the final test data. The final results are depicted in \fig{fig:G} for various choices of the loop weight $\Lambda$. We can clearly see, that the optimal choice of $\Lambda\approx1$ leads to a reduction of RMSE on six out of nine data sets. However, the relative improvement from semi-supervised learning is very small, as shown in \tab{tab:rf_ssl} and most of the time less than $1\%$. If we compare these results with other semi-supervised regression algorithms on the same data sets from \cite{wetzel2021twin}, one can observe that this improvement is significantly less than semi-supervised TNNR and slightly less than co-training with neural networks.

\section{Negative Results}
Let us briefly discuss in this section different ideas that we tried during our experiments but did not lead to a consistent improvement of twinned regression methods.

While exploring the ideal weighting of anchor during the inference phase of twinned regression methods, a straightforward idea was to try incorporating the intrinsic uncertainty metrics \cite{wetzel2020twin}. These include the ensemble standard deviation and the violation of loop consistency. Anchors with a lower uncertainty metric should be weighted higher than anchors with high uncertainty metrics. While we observed some benefit, we could not consistently show that this process improved the accuracy in a statistically significant manner. We believe that other uncertainty metrics that unrelated to the intrinsic consistency metrics might be better suited as for example in Gaussian processes.

Our initial plan when devising a strategy to adopt the semi-supervised learning framework from \cite{wetzel2021twin} to other algorithms was based on an iterative algorithm. After training the underlying twinned regression algorithm, the model would predict labels on unknown data points. Randomly sampling loops containing these data points would allow us to check for loop consistency. The unknown data points would then be added to the training data set with a label that corresponds to the original prediction slightly modified in the direction which satisfies the loop condition. The idea was to iterative refine the labels by repeating this process. However, it turned out that many times this process would either not converge for $\Lambda>1/3$, or eventually converge to sub-optimal solutions, worse than the initial supervised version.

\begin{figure}
    \centering
    \includegraphics[width=0.95\textwidth]{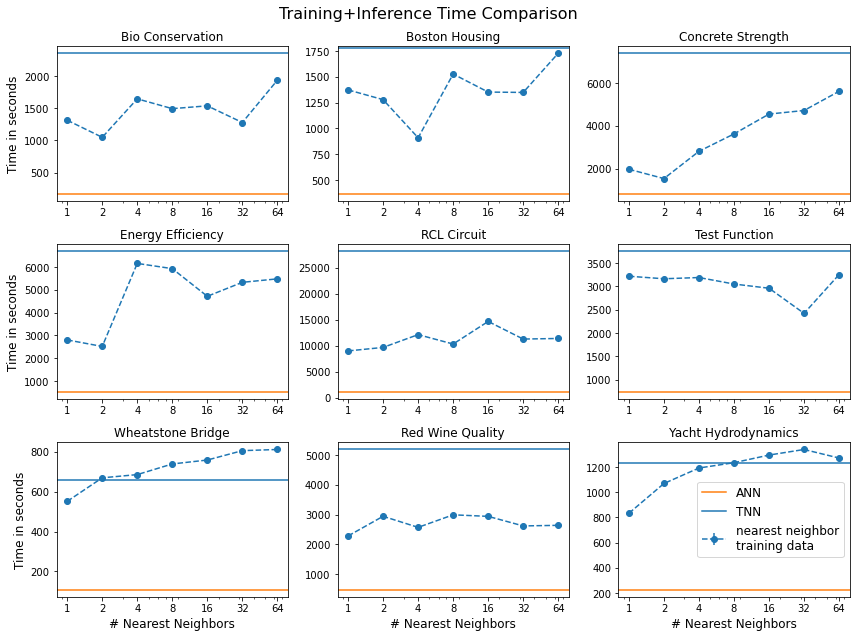}
    \caption{Training and inference time comparison in seconds between different versions of TNNR and ANN regression. The solid orange line indicates the time for training and inference in the case of ANNs. The solid blue line is the same for original TNNR. The blue dotted line indicates restricting the possible training/inference pairs to nearest neighbors, the x-axis corresponds to the number of neighbors.}
    \label{fig:C}
\end{figure}

Combining k-NN regression with TNNR was also aimed at reducing the computational time. As explored in \cite{wetzel2020twin}, the training time of twinned regression methods scales poorly towards larger data sets, mostly caused by the increase in the effective data set size through pairing of data points. While for many baseline algorithms a clear relationship between data set size and training time, for neural networks it is less known. As neural networks training time scales very favorably with training set size we focused on TNNR to test the training time improvement from only using nearest neighbor paring during training phase. In \fig{fig:C} we can see that there is a tendency for a reduced computational cost on most data sets. However, the training time scaling is too minor and inconsistent to use it as a sole justification to use NNTNNR over traditional TNNR.

\section{Conclusion}
Twinned regression is a simple and versatile framework to improve performance through intrinsic ensembling and semi-supervised learning on small to medium-sized data sets. In this article, we have answered several questions about the nature of the twinned regression framework. Further, we devised several improvements to further improve the already state-of-the-art performance of twinned regression methods.

We compared the ensemble behaviour of traditional ANNs and TNNR. For this purpose we mapped single anchor TNNR to an equivalent ANN model during inference phase. By visualizing the results of these examinations in \fig{fig:A} we can see that the performance an ensemble of single anchor TNNs converges towards an ensemble of full anchor TNNs at around 32 ensemble members. Increasing the number of anchors would not yield any additional gain. This suggests that the intrinsic ensemble diversity of TNNR is a subset of the diversity that can be achieved by retraining the network for each anchor. The combined anchor+direct ensemling containing multiple TNNs causes a much larger improvement than the ensembling of traditional ANNs as can be seen in \tab{tab:ann_ensemble} and \tab{tab:tnnr_ensemble}. This explains one element of the outperformance of TNNR over ANN regression.

Further, we examined what effect the increased training set size through pairing data points has on the TNN accuracy, \fig{fig:E}. Generally, more pairings per training data point reduced the RMSE. However, to our surprise not all data sets benefited from this pairing, on two data sets (BC,WN) TNNR had the lowest RMSE if only one pairing per training data point was allowed. By comparing this malfunction to results in \fig{fig:A} and \fig{fig:B}, we can see that it occurs in exactly the data sets where k-NN regression and ANN ensembles outperform TNNR, signaling a breakdown of the performance increasing factors of TNNR. By looking at the properties of the data sets it seems like twinned regression methods perform best on continuous data sets where the label can be approximated through a deterministic function.

We also pointed at another advantage of TNNR in the case where it is impossible to store a large number of parameters, but one wants to retain the advantages of ensembles. TNNR provides the possibility to generate an ensemble of predictions just by storing one model and some anchors, which is usually significantly smaller than storing multiple ANN models, see \fig{fig:F}.

While exploring the ideal weighting of anchors during the inference phase of TNNR, we found that nearest-neighbor predictions tend to yield the most accurate results. This lead us to develop nearest-neighbor TNNR (NNTNNR) which is a combination of the k-nearest neighbor algorithm and TNNR. There are two versions of NNTNNR, one which respects nearest neighbors during both training and inference and another version that only restricts nearest neighbors during the inference phase. Restricting to nearest-neighbor training tends to yield slightly better results \fig{fig:D}. Both versions outperform standard TNNR especially on low noise data sets, see \tab{tab:knn_tnnr}. It is important to note that NNTNNR is not just a minor improvement to k-NN regression since it has a very different performance profile when it comes to varying the number of anchors, or nearest neighbors, respectively, as can be seen in \fig{fig:B}.

We devised a semi-supervised regression framework based on enforcing loop consistency that can be applied to any twinned regression algorithm, but we tested it for random forests. This method yielded a clearly visible improvement over their supervised counterparts as can be seen in \fig{fig:G}. However, the magnitude of the reduction of the RMSE is relatively small and almost always less than $1\%$, see \tab{tab:rf_ssl}. Comparing these results with other semi-supervised regression algorithms on the same data sets from \cite{wetzel2021twin}, we can see that this improvement is significantly less than semi-supervised TNNR and slightly less than co-training with neural networks.

The code supporting this publication is available at \cite{githublink}.

\section{Acknowledgements}
Let us thank Zurab Jashi for his help with the random forest code. This work was supported by Mitacs and Homes Plus Magazine Inc. through the Mitacs Accelerate program. We also acknowledge Compute Canada for computational resources. We thank the National Research Council of Canada for their partnership with Perimeter on the PIQuIL. Research at Perimeter Institute is supported in part by the Government of Canada through the Department of Innovation, Science and Economic Development Canada and by the Province of Ontario through the Ministry of Economic Development, Job Creation and Trade.

\newpage
\appendix
\onecolumn
\section{Data sets}
\label{sec:data}
\begin{table}[h!]
  \caption{Data sets
  %\td{update with latest results}
  }
  \centering
  \begin{tabular}{lllll}

  Name & Key & Size & Features & Type\\
\midrule
  Bio Concentration & BC &779&14&Discrete, Continuous\\
  Boston Housing & BH &506 & 13&Discrete, Continuous\\
  Concrete Strength & CS &1030&8&Continuous\\
  Energy Efficiency & EF &768&8& Discrete, Continuous \\
  RCL Circuit Current &RCL&4000&6&Continuous\\
  Test Function & TF & 1000 & 2 & Continuous\\
  Red Wine Quality & WN &1599&11&Discrete, Continuous\\
  Wheatstone Bridge Voltage &WSB&200&4&Continuous\\
  Yacht Hydrodynamics & YH &308&6&Discrete\\
  \end{tabular}
\end{table}
The test function (TF) data set created from the equation
\begin{align}
F(x_1,x_2)=x_1^3+x_1^2-x_1-1+x_1x_2+\sin(x_2)
\end{align}
and zero noise.

The output in the RCL circuit current data set (RCL) is the current through an RCL circuit, modeled by the equation
\begin{align}
I_0=V_0 \cos(\omega t)/\sqrt{R^2+(\omega L-1/(\omega C))^2}
\end{align}
with added Gaussian noise of mean 0 and standard deviation 0.1.

The output of the Wheatstone Bridge voltage (WSB) is the measured voltage given by the equation
\begin{align}
V=U(R_2/(R_1+R_2)-R_3/(R_2+R_3))
\end{align}
with added Gaussian noise of mean 0 and standard deviation 0.1.

\section{Bias-Variance Tradeoff and Ensembles}
\label{app:bv}
In a regression problem, one is tasked with finding the true labels on yet unlabelled data points through the estimation of a function $f(x)=y$. Given a finite training data set $D$ we denote this approximation $\hat{f}(x;D)$. The expected mean squared error can be decomposed by three sources of error, bias error $\operatorname{Bias}_D[\hat{f}(x;D)]$ , variance error $\operatorname{Var}_D\big[\hat{f}(x;D)]$ and intrinsic error of the data set $\sigma$. 
\begin{align}
\text{MSE} = \operatorname{E}_x\bigg\{\operatorname{Bias}_D[\hat{f}(x;D)]^2+\operatorname{Var}_D\big[\hat{f}(x;D)\big]\bigg\} + \sigma^2.
\end{align}
If we replace the estimator by an ensemble of two functions $\hat{f}(x;D)=1/2\hat{f}_A(x;D)+1/2\hat{f}_B(x;D)$, each exhibiting the same bias and variance as the original estimator, then we can decompose the MSE
\begin{align}
\text{MSE} &= \operatorname{E}_x\bigg\{\operatorname{Bias}_D[1/2\hat{f}_A(x;D)+1/2\hat{f}_B(x;D)]^2+\operatorname{Var}_D\big[1/2\hat{f}_A(x;D)+1/2\hat{f}_B(x;D)\big]\bigg\} + \sigma^2 \\
&= \operatorname{E}_x\bigg\{\operatorname{Bias}_D[\hat{f}(x;D)]^2+\operatorname{Var}_D\big[1/2\hat{f}_A(x;D)\big]+\operatorname{Var}_D\big[1/2\hat{f}_B(x;D)\big]\\
&\hspace{5cm}+2\operatorname{Cov}_D\big[1/2\hat{f}_A(x;D),1/2\hat{f}_B(x;D)\big]\bigg\} + \sigma^2 \\
&= \operatorname{E}_x\bigg\{\operatorname{Bias}_D[\hat{f}(x;D)]^2+1/2\operatorname{Var}_D\big[\hat{f}_A(x;D)\big]+1/2\operatorname{Cov}_D\big[\hat{f}_A(x;D),\hat{f}_B(x;D)\big]\bigg\} + \sigma^2
\end{align}
The more uncorrelated $\hat{f}_A(x;D)$ and $\hat{f}_B(x;D)$ are, the smaller is the ratio between variance and covariance. Thus an ensemble consisting of weakly correlated ensemble members reduce the MSE by circumventing the bias-variance tradeoff. By induction this argument extends to larger ensemble sizes.

\section{Neural Network Architectures}
\label{sec:nn_architecture}
Both our traditional neural network regression and twin neural network regression methods are build using the same architecture build using the tensorflow library \cite{Tensorflow2015}. They consist of two hidden layers with 128 neurons each and relu activation functions. The  final layer contains one single neuron without an activation function. We train our neural networks using the adadelta optimizer, and use learning rate and early stop callbacks that reduce the learning rate by 50\% or stop training if the loss stops decreasing. For this reason it is enough to set the number of epochs large enough such that the early stopping is always triggered, in our cases this is 2000 for ANNs and 10000 for TNNR. The batchsizes are in both cases 16.

\section{Random Forests}
\label{sec:rf}
The random forests in this article use the scikit-learn library \cite{scikit-learn}. They are trained on a subset of all data sets: from each data set, we randomly sample 100 training data points and 100 test data points. We use five-fold cross-validation to optimize the following hyper-parameters of our random forests:
    'max\_depth' $\in [ 4, 8, 16, 32, 64]$, 
    'max\_features' $\in [0.33, 0.667, 1.0]$, 
    'min\_samples\_leaf' $\in[1, 2, 5]$, 
    'min\_samples\_split' $\in  [2, 4, 8]$, 
    'n\_estimators' $\in  [100, 300, 600]$.
Both, the traditional and the twinned random forests choose their optimal hyper-parameters from the same pool. It is important to note that for semi-supervised learning the hyper-parameters are only optimized during the initial supervised learning step, the optimal parameters are then carried forward to be used during semi-supervised learning.
\begin{table}
\centering
  \caption{Best estimates for test RMSEs obtained by Random Forest compared to Twinned Random Forests
  }
 
  \begin{tabular}{l llr}\\
    %%%%%%%   & \multicolumn{3}{l}{80\% labelled training data}            \\

 & Random Forest & Twinned Random Forest & Improvement \\
   \cmidrule(r){2-2}\cmidrule(r){3-4}
Bio Conservation & 0.7407$\pm$ 0.0087 & 0.7427$\pm$ 0.0081 & -0.27\% \\
Boston Housing & 4.0019$\pm$ 0.1394 & 3.8301$\pm$ 0.1322 & 4.29\% \\
Concrete Strength & 6.3763$\pm$ 0.2136 & 5.8519$\pm$ 0.2242 & 8.22\% \\
Energy Efficiency & 1.8773$\pm$ 0.0422 & 1.8906$\pm$ 0.0417 & -0.71\% \\
RCL Circuit & 0.3168$\pm$ 0.0071 & 0.2958$\pm$ 0.0068 & 6.63\% \\
Test Function & 0.1402$\pm$ 0.0044 & 0.0874$\pm$ 0.0023 & 37.66\% \\
Wheatstone Bridge & 0.1461$\pm$ 0.0039 & 0.0942$\pm$ 0.0025 & 35.52\% \\
Red Wine Quality & 0.5989$\pm$ 0.0085 & 0.6068$\pm$ 0.0083 & -1.32\% \\
Yacht Hydrodynamics & 1.1917$\pm$ 0.029 & 1.1117$\pm$ 0.0383 & 6.71\%

  \end{tabular}
  \label{tab:twin_rf}
\end{table}

\begin{algorithm}
\caption{Semi-Supervised Learning through Loop Consistency}\label{alg:semi_sup}
  \KwData{Labelled data set $D_L=(X_{train,L},Y_{train,L})$\\ \ Unlabelled data set $D_U=(X_{train,U})$}
  \KwInput{Loop weight $\Lambda$\\Loop number $n_l=$length$(D_L)/3$}

  create $\tilde{D}_L=[((x_i,x_j,x_i-x_j),y_{ij}=y_i-y_j) $ for $ x_i \in X_{train,L} $ for $ x_j \in X_{train,L}]$\\
  initialize machine learning model $M$\\
  train $M$ on $\tilde{D}_L$\\

 sample $n_l$ loops L=[$(x_i,x_j,x_k)$ where $x_i \in X_{train,L}$, $x_j,x_k \in X_{train,U}$]\\
  \For{$(x_i,x_j,x_k) \in L$}{
  
  predict $M(x_i,x_j),M(x_j,x_k),M(x_k,x_i)$\\
  $a=M(x_i,x_j)+M(x_j,x_k)+M(x_k,x_i)$\\

  $(y_{ij}, y_{jk},y_{ki})=(M(x_i,x_j)-\Lambda a,M(x_j,x_k)-\Lambda a,M(x_k,x_i)-\Lambda a)$\\

 add $((x_i,x_j,x_i-x_j),y_{ij})$,$((x_j,x_k,x_j-x_k),y_{jk})$,$((x_k,x_i,x_k-x_i),y_{ki})$ to $\tilde{D}_L$}
  train $M$ on $\tilde{D}_L$\\

  \KwOutput{Trained Model $M$}

\end{algorithm}

\newpage
\phantom{a}

\bibliography{library}
\bibliographystyle{iopart-num}
\end{document}